# Automatic landmark annotation and dense correspondence registration for 3D human facial images

**Jianya Guo, Xi Mei, Kun Tang***

CAS-MPG Partner Institute and Key Laboratory for Computational Biology, Shanghai Institutes for Biological Sciences, Chinese Academy of Sciences, Shanghai, China.

**Funding:** This work was supported by the Cutting Edge Research Project (Grant No. 2011KIP201) from the CAS-SIBS Outstanding Junior Researchers, the Key Research Direction Grant (No. KSCX2-EW-Q-1-12) from the CAS Knowledge Innovation Project and the Max-Planck-Gesellschaft Partner Group Grant. The funders had no role in study design, data collection or analysis, decision to publish, or preparation of the manuscript.

**Competing interests:** The authors have declared that no conflicts of interest exist.

*corresponding author:

Kun Tang          Telephone: 86-21-54920277 Fax: 86-21-54920451

Email: tangkun@picb.ac.cn

# Abstract

Dense surface registration of three-dimensional (3D) human facial images holds great potential for studies of human trait diversity, disease genetics, and forensics. Non-rigid registration is particularly useful for establishing dense anatomical correspondences between faces. Here we describe a novel non-rigid registration method for fully automatic 3D facial image mapping. This method comprises two steps: first, seventeen facial landmarks are automatically annotated, mainly via PCA-based feature recognition following 3D-to-2D data transformation. Second, an efficient thin-plate spline (TPS) protocol is used to establish the dense anatomical correspondence between facial images, under the guidance of the predefined landmarks. We demonstrate that this method is robust and highly accurate, even for different ethnicities. The average face is calculated for individuals of Han Chinese and Uyghur origins. While fully automatic and computationally efficient, this method enables high-throughput analysis of human facial feature variation.

**Key Words:** 3D face, facial morphology, registration, landmark localization, dense correspondence.

# Introduction

Large-scale, high-throughput phenotyping is becoming increasingly important in the post-genomics era. Advanced image processing technologies are used more and more for collecting deep and comprehensive morphological data from different organisms, such as yeast [1], plants [2], and mice [3]. The soft tissue of the human face is a complex geometric surface composed of many important organs, including eyes, nose, ears, mouth, etc. Given its essential biological functions, the human face has been a key research subject in a wide range of fields including anthropology [4], medical genetics [5,6,7,8], forensics [9,10], psychology [11,12], and aging [13,14]. With the development of non-invasive 3D image acquiring technologies, high resolution 3D data of the human face are becoming readily available for various applications (www.3dmd.com).

In many research fields, only a small fraction of the high-resolution 3D image data is used, which usually comprises a set of landmarks and/or their mutual distances and angles [4,15,16]. However, methods have been developed to register the 3D surfaces using their dense surface meshes, which allows the inclusion of the complete data set for powerful inferences and analyses [8,17,18]. In general, surface registration methods can be classified into two groups: rigid and non-rigid techniques. The former aligns surfaces by rigid transformation, e.g, rotation and translation, while the latter employs deformations to get a close alignment between surfaces. For rigid registration, the Iterative Closest Point (ICP) algorithm is the most widely used [19]. However, as the affine transformations do not fully capture the anatomical variability, the closest corresponding points between two surfaces calculated by the ICP algorithm are not necessarily biologically homologous, especially when comparing faces which differ significantly. In order to establish the anatomical correspondence more effectively, it is necessary to employ non-rigid transformations. A common method for deforming 3D surfaces is the thin-plate spline (TPS) algorithm [20]. The process of using TPS warping involves minimizing a bending energy function for a transformation over a set of fiducial points (landmarks), thereby bringing the corresponding fiducial points on each surface into alignment with each other.

Although it is a powerful method, TPS has a key drawback that limits its use in the analysis of large-scale, open 3D facial datasets: namely, it requires a set of landmarks to be annotated first. For many existing registration methods, landmarks have to be manually labeled on the facial surfaces [21,22,23,24], which is highly time consuming and introduces human errors. Methods have been developed to combine ICP-based landmark annotation and TPS warping to automate the registration [25,26]. However, the landmark correspondences found by ICP are not exactly anatomically homologous, as previously discussed. There exist many automatic landmark localization methods [9,27,28,29,30,31]. At some point, most of these approaches use local, curvature-based facial features due to their invariance to surface translation and rotation. The two most frequently adopted features are the HK curvature and the shape index [27,28,29,32]. However, curvature-based descriptors often suffer from surface irregularities, especially near eye and mouth corners [33]. Other studies have used appearance-based methods where the facial features are modeled by basis vectors calculated from transformations such as Principal Component Analysis (PCA) [34,35], Gabor wavelets [36,37], or the Discrete Cosine Transform (DCT) [31].

In this study, we develop an automatic registration method which combines a novel workflow of landmark localization and an efficient protocol of TPS-based surface registration. The landmark localization mainly employs PCA to extract landmarks on surfaces by use of both shape and texture information. For the surface registration, a new TPS warping protocol that avoids the complication of inverse TPS warping (a compulsory procedure in the conventional registration method) is used to resample the meshes according to the reference mesh. We show that this method is highly accurate and robust accross different ethnicities. We also propose a new spherical resampling algorithm for re-meshing surfaces which efficiently removes the caveats and improves the mesh structure. Furthermore, the associated texture is also included in the registered data for visualization and various analyses.

## Materials and Methods

### Ethics Statement

Sample collection in this study was carried out in accordance with the ethical standards of the ethics committee of the Shanghai Institutes for Biological Sciences (SIBS) and the Declaration of Helsinki, and has been specifically surveyed and approved by SIBS. A written statement of informed consent was obtained from every participant, with his/her authorizing signature. The participants, whose transformed facial images are used in this study as necessary illustrations of our methodology, have been shown the manuscript and corresponding figures. Aside from the informed consent for data sampling, a consent of publication was shown and explained to each participant and their authorizing signature was obtained as well.

**The 3D face data set**

Three-dimensional facial images were acquired from individuals of age 18 to 28 years old, among which 316 (114 males and 202 females) were Uyghurs from Urumqi, China and 684 (363 males and 321 females) were Han Chinese from Taizhou, Jiangsu Province, China. Another training set which did not overlap with the first 1000 sample faces, consisted of 80 Han Chinese, 40 males and 40 females from Taizhou, Jiangsu Province, China. The participants were asked to pose an approximately neutral facial expression, and the 3D pictures were taken by the 3dMDface® system (www.3dmd.com). Each facial surface was represented by a triangulated, dense mesh consisting of ~30000 vertices, with associated texture (figure 1).

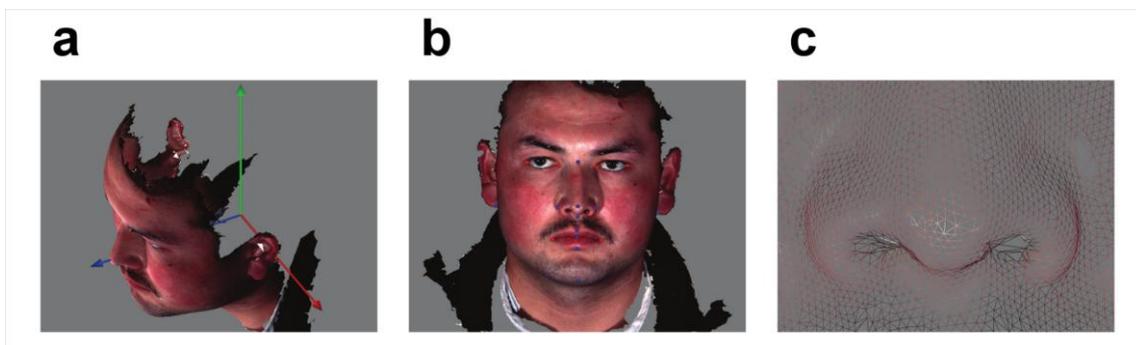

**Figure 1.** The surface used in our research. **a.** The coordinate system used in our research (red, green and blue axes stand for x, y and z axes respectively). **b.** An example scan with 17 landmarks marked by the colored spots. The red spots are the 6 most salient landmarks, namely the inner and outer corners of the eyes and both corners of the mouth, the blue spots indicate the other 11 landmarks used in this study. **c.** Raw mesh details around the nose tip.

**Workflow**

The workflow is briefly described as follows (figure 2). Starting with a set of raw 3D face scans, the nose tip is first automatically localized on each face using a sphere fitting approach and pose normalization is performed to align all sample faces to a uniform frontal view. For the landmark annotation, the six most salient landmarks were first manually labeled on a set of training samples; Principal Component Analysis (PCA) was then employed to localize these 6 landmarks on the sample surfaces and 11 additional landmarks were heuristically annotated afterwards. A reference face was then chosen, re-meshed using spherical sampling, and TPS-warped to each sample face using the 17 landmarks asfiducial points. A dense, biological correspondence was thus built by re-meshing the sample face according to the reference face. The correspondence is further improved by using the average face as the reference and repeating the registration process.

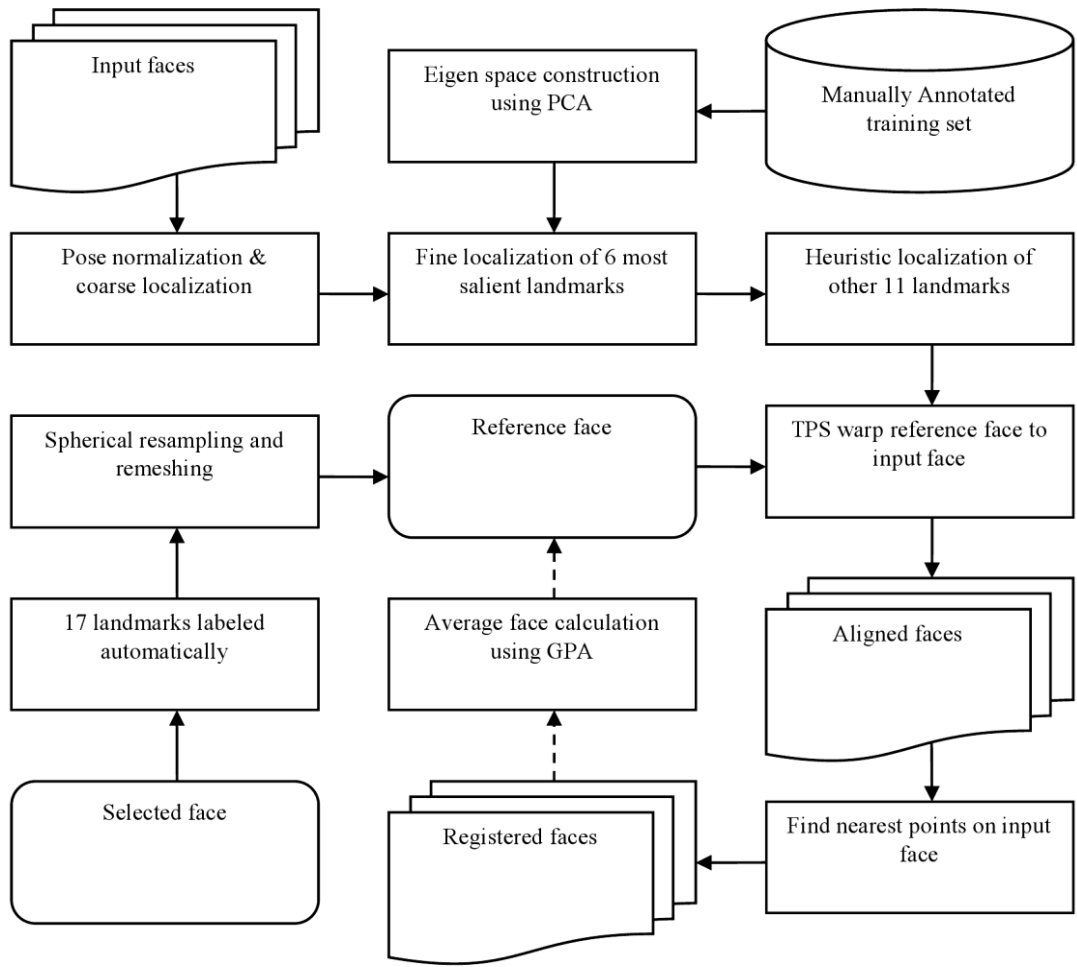

**Figure 2.** The workflow of the analysis

**Preliminary nose tip localization and pose normalization**

In 3D facial image processing, pose normalization and landmark localization are highly dependent on each other since pose normalization is typically guided by landmarks. The features commonly used for pose correction are the nose tip and inner eye corners as they are easier to detect [27], less sensitive to pose variation, and invariant to facial expressions [32,34,38,39]. On the other hand, most existing landmark localization approaches rely on the assumption of frontal or approximately frontal poses and are therefore sensitive to roll and yaw rotation [30,32,40]. In order to fully automate the pose normalization and landmark annotation, we first identify the most robust and prominent landmark, the nose tip.

Since the area around the nose tip can be approximated as a semi-sphere with a diameter specific to nose, we try to identify the nose tip by fitting a sphere around every vertex using its surrounding vertices. A vertex is likely the nose tip if its neighboring points fit a sphere very well and the sphere diameter approaches the specific value of the nose tip. As this method is insensitive to the pose of the face, the spherical area on nose tip can be seen as a rotation invariant descriptor (RID). The algorithm is described as follows. Let us denote a facial mesh composed of $N$ points by $\mathbf{F} = \{\mathbf{p}_i\}$ for $i = 1, \ldots, N$. Suppose $\mathbf{S}$ is the set of $M$ points that are within distance $R$ around the point $\mathbf{p_j}$ ($1 \leq j \leq N$). The best fit sphere $\mathbf{T}$ around $\mathbf{p_j}$ is therefore determined by two parameters, namely the center $\mathbf{O} = (a, b, c)$ and radius $r$. Another parameter $e$ is the average residual distance of the $M$ points to the best fit sphere. $e$ describes how well the set of $M$ points fit onto a sphere. A detailed description of sphere fitting and the calculation of $e$ can be found in appendix I. The smaller $e$ is, the better $\mathbf{S}$ fits a sphere. The two parameters, $r$ and $e$, are calculated for every point. In order to form a proper sphere of radius $r$ around each vertex, the included distance to adjacent points ($R$) must be slightly larger than the radius of the sphere ($r$) as it is assumed that not every point will lie on the sphere. On the other hand, $r$ should be chosen with good consistency across genders and ethnic backgrounds, thereby establishing a uniform criterion for all faces. From a large number of experiments, we determined a typical $r$ value, hereafter denoted as $r_0$ that showed little variance across the four test groups of different sex and ethnicities.

The two spherical parameters can then be combined with the optimal $r_0$ radius into one statistic ($f$) which describes how well a given point fits the criteria for a nose tip:

$$f = e(r_0 + |r - r_0|). \tag{1}$$

The $f$ value should be very small around the nose tip region. Indeed, we found that small $f$ values congregated around the nose tip area (figure 3). More interestingly, the global minima of the $f$ values consistently appeared close to the manually annotated nose tip across hundreds of observations. We therefore use the point with the minimum $f$ value to approximate the nose tip (figure 3).

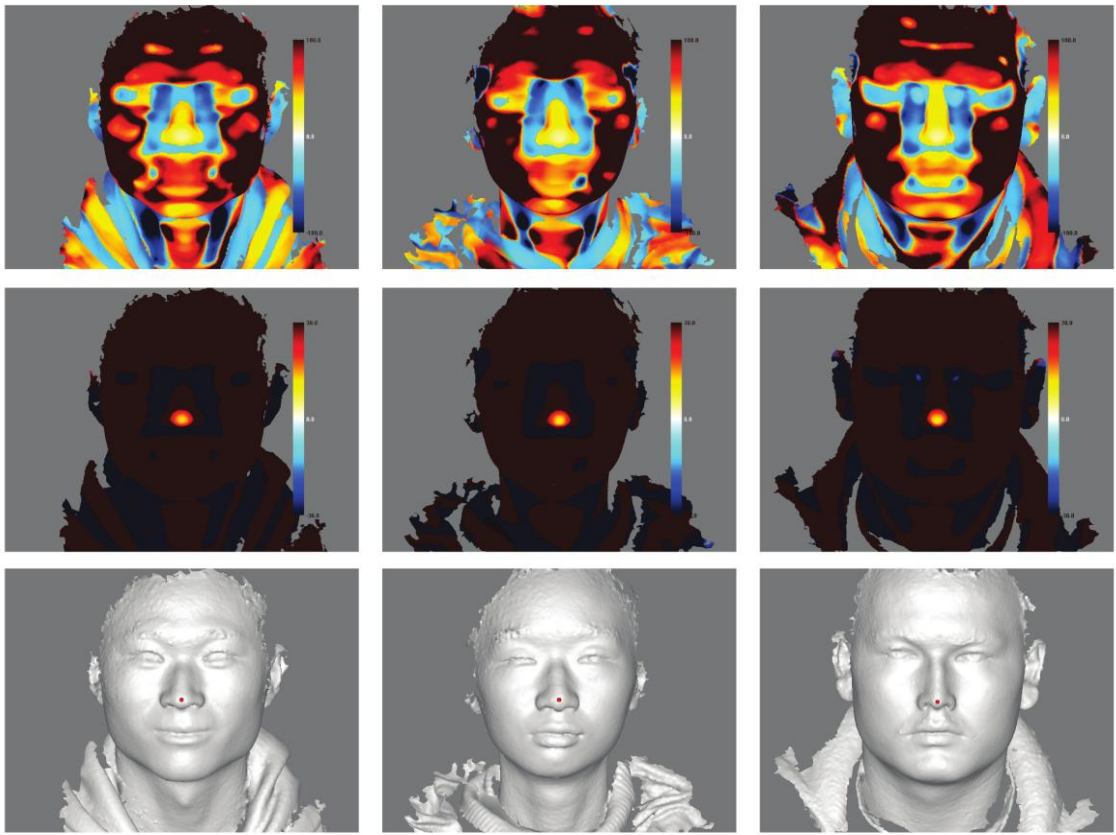

**Figure 3.** Nose tip localization using RID illustrated for three individuals: a Han Chinese female in the left column, a Han Chinese male in the middle column and an Uyghur female in the right column. Top row, the $f$ values are shown as color gradients. Warm colors indicate convex sphere fitting,

while the cold colors indicate concave to the reader. The *f* values deviating more from 0 are marked with greater color intensity. Central row, the minimum convex *f* values plotted for different individuals, which can be seen to coincide with the manually annotated nose tips shown in the bottom row.

The pose correction becomes easy once the nose tip has been located. Correcting the pose basically consists of resetting the viewing coordinate system where an origin point and two axes must be defined. In some studies, the ICP matches are applied [41,42]. Other studies try to find landmarks (i.e. inner eye corners) other than the nose tip to determine the pose [20,30]. However, in this study we followed a rather practical solution in which all vertices within 50*mm* of the nose tip are used to correct the pose via the Hotelling transformation [41,42].

**Localization of the six most salient landmarks using PCA**

Here we propose a novel landmark localization method. The basic idea is to transform the 3D shape and texture data into a 2D space. A 2D PCA algorithm is then used to identify the six most salient landmarks, namely the inner and outer corners of the eyes and both corners of the mouth (figure 1b).

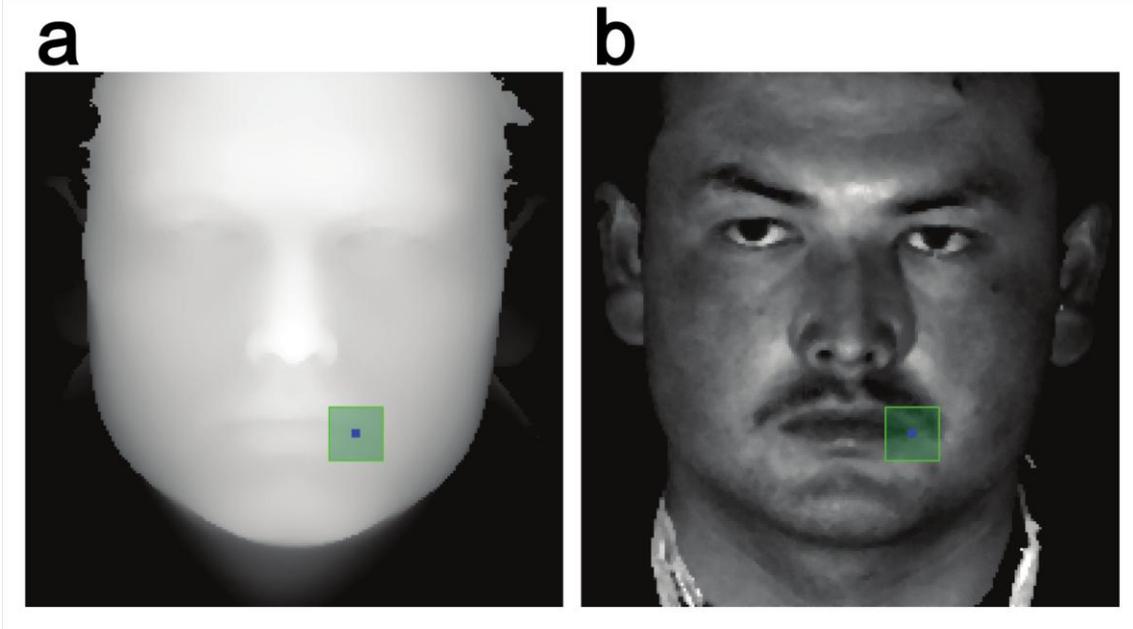

**Figure 4.** The signature patch for the left lip corner illustrated in the 2D space. **a.** The z coordinate values mapped into the 2D space. **b.** The gray scale values mapped into the 2D space.

First, the image texture is converted to the YCbCr color space, in which the Y component defining the gray scale intensity is calculated as $y = 0.30r + 0.59g + 0.11b$. Only the gray scale values are used as color information for this step. For any 3D face image, the plane defined by the x and y axes is defined as the target 2D space. The 3D surface and its corresponding 2D texture are then resampled on a uniform square grid at a 1*mm* resolution to obtain the corresponding z coordinate values and gray scale values. These values are directly mapped to the target 2D space (figure 4). In order to minimize the data noise, the z coordinate and gray scale values are de-noised using a $3 \times 3$ decision-based median filter [29]. Only the values of the outer most layer are transformed to 2D following the z buffering algorithm, particularly for the areas where the 3D surface folds into a multilayer along the z-axis [43]. Holes that may occur inside the surface are closed by bicubic interpolation as previously described [44]. The interpolation process was done separately on texture and the 2.5D image data. The resulting 2D image combines both shape and texture information, which serves as the basis for the PCA-based landmark localization. The PCA analysis is a commonly used approach for accurate pattern recognition in 2D data [24,43,45]. It involves retrieving the feature signatures by dissecting the training data with PCA, followed by projecting the sample data into the PCA eigenspace to determine the similarity. In this study, the landmark signature is obtained by defining a patch of a given size, say $s\ mm \times s\ mm$, centered around the manually annotated landmark in the training set (figure 4). Each patch therefore contains $s^2$ z coordinate values, which are then concatenated into a vector and normalized to have zero mean and unit length. We define it as $Z = (z_1, z_2, ..., z_{441})$. The same number of gray scale values are also concatenated into a vector and normalized to have unit length. We define it as $Y = (y_1, y_1, ..., y_{441})$. $Z$ and $Y$ can be combined together to specify the shape and texture properties around the landmark:

$$P = (z_1, y_1, z_2, y_2, \cdots, z_{441}, y_{441})^T. \tag{2}$$

$P$ is then calculated for the signature eigenspace **U** using PCA (see Appendix II for details). To find the landmarks in a sample face, a patch of $s\ mm \times s\ mm$ is similarly defined for every point in the corresponding 2D grid, and a sample patch vector $P_s$ is derived following equation (2). $P_s$ is subsequently projected to the space **U** to evaluate its closeness to the origin point of **U**. In this study, two measurements of closeness are used, the reconstruction error $e$ and the Mahalanobis distance $d$ (see Appendix II for details). Sample points with smaller values for $e$ and $d$ are more likely to be a valid landmark. Therefore, the sample point corresponding to the minimum product value of *e* and *d* is defined as the landmark in our work. The patch size parameter *s* inevitably affects the final localization accuracy. We found that in general, the localization error decreases with the increasing patch size. The *s* values of 16 – 30 *mm* all seem to result in proper localization. In this study we chose a s value for which the error approaches minimum while the patch size is relatively small. To further optimize the computational efficiency, we narrow down the search for each landmark to a corresponding "landmark zone" on each sample face. Briefly, an arithmetic mean is calculated for each landmark across the training set, and projected onto the 2D space. Rectangular areas around the projection points are then defined as the landmark zones, with their sizes set experimentally (i.e. by training through a large number of faces) to ensure all real landmarks are encompassed. Therefore, the search for a particular landmark is done only within this landmark zone.

**Heuristic localization of ten additional landmarks**

Given the annotation of the six most salient landmarks, the pose of the surface can be fine tuned again. The reference plane is set to be the best fit plane to the six landmarks by least squares. The normal to the reference plane is set to be the z axis, and the y axis is given by the projection of the line going through the centers of lip corners and the eye corners onto the reference plane. The x axis is uniquely determined afterwards.

After the pose correction, 10 additional landmarks are identified heuristically by using geometric relations and texture constraints and the nose tip position is also updated. These 1andmarks include soft tissue nasion, alares, subnasale, labiale superius (upper lip point), stomion (the middle point between the upper and lower lip), labiale inferius (lower lip point), pogonion (chin point), and earlobe tips. The nose tip can be fine tuned according to the more uniformly defined coordinate system across all sample surfaces. Briefly, a semi-sphere is refitted around the previous nose tip and the point that minimizes the z coordinate error is chosen as the new nose tip. The subnasale point can be located by finding the inflection point with the minimum angle right below the nose tip. The alare points are the inflection points with the minimum local angles going horizontally away from the nose tip. Similar angle heuristics are applied to the detection of labiale superius, inferius, and stomion, with additional texture information in the YCbCr color space. For example, the labiale superius should locate the position on the border line where the Cr values below the line are greater (more red) than those above. Noticing that the region around the nasion point is approximately saddle-shaped and that of the chin point is ellipsoidal or sphere-shaped, both characterized by the two-way symmetry, we therefore locate the two points by finding the maximum local symmetry scores. The earlobe points are easily found by locating the tips with sheer slopes along the z-axis.

**Spherical resampling and surface remeshing**

During the 3D image acquisition, the surface meshes often suffer from mesh structure irregularities and/or defects such as mesh holes (see figure 1c for example). Surface remeshing is often used to solve such problems [46]. In this work, we apply spherical sampling to the reference surface to obtain a well-structured mesh. Spherical sampling is preferred as human faces are approximately ellipsoidal. We first perform a spherical parameterization to the surface using the geographic coordinates. Given a vertex $(x_i, y_i, z_i)$ on the original surface mesh, the spherical parameterization $(\rho_i, \theta_i, \varphi_i)$ can be obtained as follows:

$$\begin{aligned}\rho_i &= \sqrt{2x_i^2 + y_i^2 + z_i^2} \\ \theta_i &= \arcsin(y_i / \rho_i) \\ \varphi_i &= \operatorname{arctanx}(\sqrt{2} x_i / z_i)\end{aligned} \qquad (1)$$

The x-coordinate is multiplied by a factor $\sqrt{2}$ before the coordinate conversion, to compensate for the face aspect ratio (height to width) [42]. When plotted against $\theta$ and $\varphi$, the parameterized surface unfold into a nearly flat plane. This surface is then trimmed with an oval path to remove the irregular edges and re-sampled from a uniform square grid with an interval of 0.005 for both $\theta$ and $\varphi$. The re-sampled data points are then converted back to the Cartesian coordinate system to define a new surface mesh.

**Surface registration for dense correspondence**

In order to preserve the anatomical correspondence across the facial surfaces, we adopted the idea of the TPS-based registration method proposed previously [23]. In that study, all surfaces were first manually annotated for a set of landmarks. The sample surfaces and the reference were all TPS warped to the cross-sample average landmarks. Each sample surface was then re-meshed by the closest points to the reference vertices, and further inverse TPS warped back to the original shape. Mathematically, TPS warping is not invertible. Although an approximation exists, it is computationally intensive and error prone [47]. In our study, we designed an alternative scheme. First, a well-structured surface with few defects is chosen as the reference face, and spherically remeshed as described above. Then only the reference surface is TPS warped to each sample surface, taking the 17 landmarks as the fiducial points. The TPS warping is done as previously described [48]. Thereafter the vertices on the reference surface find their closest projections on the sample surface, which define the new mesh vertices of the sample surface [48,49]. The dense correspondence is established after all the sample surfaces are remeshed using the same reference. This approach eliminates the need for inverse TPS warping, and enhances the computational efficiency as well.

## Results:

**Accuracy of the landmark localization**

In this section we demonstrate the accuracy of the proposed algorithm for automatic landmark localization. The accuracy is measured by the deviation of the automatically annotated landmarks from those manually annotated.

A subset of the sample surfaces were picked randomly and manually annotated with the 17 landmarks by the experimenter who did the same to the training set. Automatic landmark annotation was also performed independently. The surfaces missing some features such as the earlobes were removed from further analysis. This left 115 Han Chinese (56 males, 59 females) and 124 (48 males, 76 females) Uyghur for the evaluation. The mean and standard deviation (SD) of the annotation errors measured in Euclidean distance, as well as the root mean square (RMS) errors were calculated (Table 1). As can be seen from table 1, most landmarks have mean errors between 1*mm* and 1.5*mm*, indicating rather high accuracy. Most of the SD values are below 1*mm*, suggesting good consistency of the annotation across different samples. The RMS error is within the range of 1.1~2*mm* for most measurements. Greater errors are found for the Pogonion (~1.8*mm* mean error for both the Han Chinese and Uyghur) and the two earlobe tips (mean error 2~3*mm*, SD error 1.6~2.2*mm* and RMS error 2.6~3.6*mm*). Pogonion and earlobes are both strongly affected by facial/head hair, which may account for the relatively larger errors and standard deviations. It is worth noticing that all the error values are similar between the Han Chinese and Uyghur samples despite the use of the Han Chinese training set. Given the substantial genetic and morphological differences between these two ethnic populations, this indicates good robustness of our method when applied to different ethnicities.

**Table 1.** Mean error and standard deviation of 17 automatically annotated landmarks.

| Landmarks | Han | | | Uyghur | | |
|---|---|---|---|---|---|---|
| | Mean Error (mm) | SD error (mm) | RMS error | Mean Error (mm) | SD error (mm) | RMS error |
| Right Eye Outer Corner | 1.339 | 0.947 | 1.641 | 1.511 | 1.091 | 1.864 |
| Right Eye Inner Corner | 1.192 | 1.155 | 1.660 | 1.280 | 0.914 | 1.573 |
| Left Eye Inner Corner | 1.162 | 0.809 | 1.416 | 1.489 | 0.896 | 1.738 |
| Left Eye Outer Corner | 1.148 | 0.738 | 1.365 | 1.507 | 1.008 | 1.814 |
| Right Lip Corner | 0.995 | 0.581 | 1.153 | 1.424 | 0.817 | 1.642 |
| Left Lip Corner | 1.012 | 0.569 | 1.161 | 1.147 | 0.693 | 1.341 |
| Nose Tip | 0.761 | 0.747 | 1.067 | 1.113 | 0.549 | 1.242 |
| Nasion | 1.487 | 0.654 | 1.625 | 1.604 | 0.833 | 1.808 |
| Right Alare | 1.310 | 0.752 | 1.511 | 1.034 | 0.742 | 1.273 |
| Left Alare | 1.480 | 0.798 | 1.682 | 1.128 | 0.577 | 1.268 |
| Lip Center | 1.189 | 0.715 | 1.388 | 1.145 | 0.812 | 1.404 |
| Upper Lip | 1.270 | 0.769 | 1.485 | 1.727 | 1.138 | 2.069 |
| Lower Lip | 1.380 | 0.855 | 1.624 | 1.501 | 0.941 | 1.772 |
| Subnasale | 1.299 | 1.263 | 1.812 | 0.999 | 0.574 | 1.153 |
| Pogonion | 1.809 | 1.878 | 2.608 | 1.785 | 0.882 | 1.992 |
| Right Earlobe tip | 2.074 | 1.658 | 2.656 | 2.866 | 2.18 | 3.601 |
| Left Earlobe tip | 2.678 | 1.640 | 3.141 | 2.835 | 2.054 | 3.501 |

**Robustness of the registration method**

One way to evaluate the robustness of the registration method is to determine how the use of different references would affect the correspondence mapping. We performed such an evaluation, as shown in figure 5. First, we obtained the average Han Chinese male and female faces by registering all the Han Chinese samples to the same reference surface, followed by obtaining the average meshes across either gender group (average face calculation is explained in more detail in the next section). These average faces are point-to-point corresponded. We can see the two average faces differ substantially in their shape (figure 5a and 5c). To test the robustness of the registration method, a test face (figure 5b) is chosen randomly from the data set, and registered separately using either average face as the reference. Euclidian distances are calculated for the pairing points between the two newly registered meshes. One expects to see small differences between the two registration results if the method is robust. Figure 5d shows that most parts have point-wise errors much less than 0.9*mm*, which indicates the robustness of our registration method with varying references. Certain regions like eyebrows exhibit greater errors, most likely due to the mesh irregularities caused by facial hair.

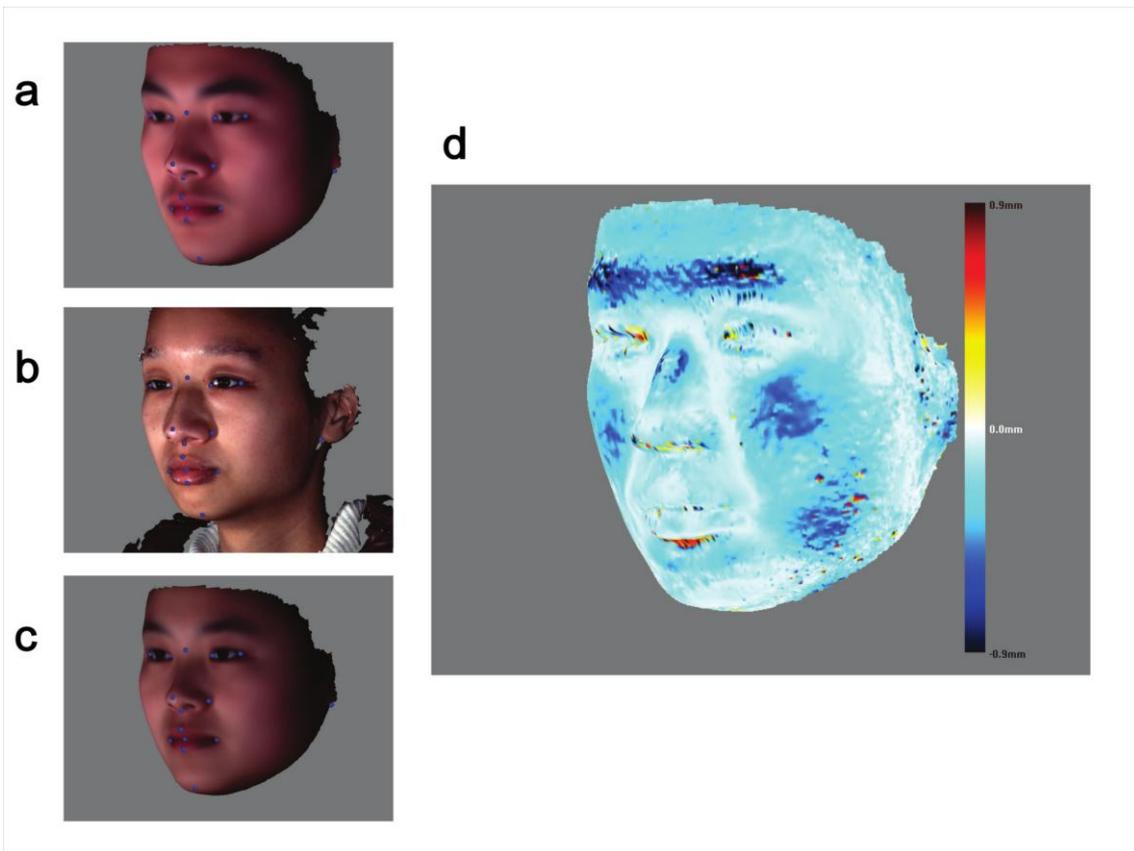

**Figure 5.** Evaluation of the robustness of the registration method. The average face of either gender is used as the reference to register the sample surface, and the registration results are compared. **a.** The average face of male Han Chinese, **b.** The sample face to be registered, **c.** The average face of female Han Chinese, **d.** The comparison of the two registration results. The differences are represented in color gradients, with the darker colors denoting greater pointwise differences.

## The average faces calculation with the 3D face registration

We applied the proposed 3D face registration method to the whole 3D face sample set. In total 363 male and 321 female Han Chinese and 114 male and 202 female Uyghur were included in this analysis. All surfaces were automatically annotated. One Han Chinese face with few caveats and fully exposed skin was chosen as the reference, to which all the sample faces were registered. The Generalized Procrustes Analysis (GPA) was then used to align all the registered surfaces to a common coordinate [50]. The average faces were then calculated as the average meshes colored by the corresponding average texture pixels across all the samples in each group. Figure 6 shows the average faces of the four groups. As can be seen, the average faces well retain the morphological and textural features of each group. Specifically, the Uyghur are characterized by a more protruding nose and eyebrow ridges while Han Chinese have wider cheeks. The skin pigmentation also seems lighter for the Uyghur compared to the Han Chinese. This difference could not be confirmed as the environmental light was not well controlled when the 3D images were taken.

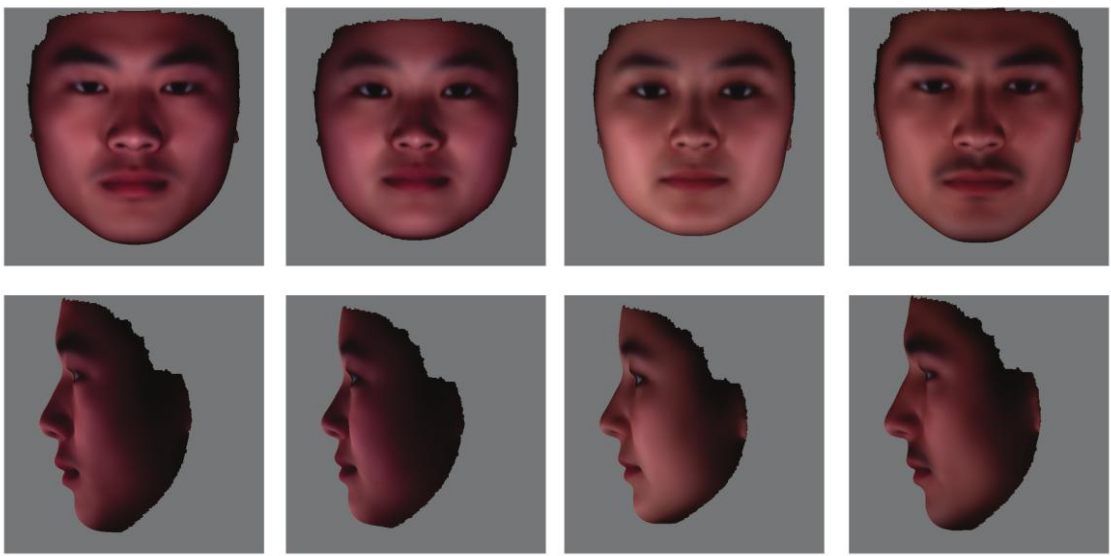

**Figure 6.** Front and profile views of the average faces. From left to right: Han Chinese male, Han Chinese Female, Uyghur female, and Uyghur male.

## Discussion:

In this work we propose a fully automatic registration method for high resolution 3D

facial images. This method combines automatic landmark annotation and TPS-based registration. Up to now, there have been few studies that have fully automated the entire dense 3D facial surface registration process. Methods have been developed to combine the ICP-based landmark localization with the TPS-based registration [25,26]. Nonetheless, ICP may fail to detect biological correspondence when surfaces differ greatly in shape [26]. Most time-honored solutions for landmark localization deal with only 2.5D data, leaving out the texture information. In particular, Perakis et al. described a method that made use of a comprehensive list of local shape descriptors, and achieved a precision of around 4mm [51]. Szeptycki et al. combined curvature analysis with a generic face model in a coarse-to-fine workflow, which enabled rotation invariant 3D landmark annotation at a precision of around 10mm [29]. On the other hand, D'Hose et al. made use of the Gabor wavelets to extract curvature information for coarse landmark localization, followed by an ICP-based fine mapping [37]. This study achieved an overall precision level of a bit over 3mm [37]. Hutton et al. developed a method called the dense surface model (DSM), which hybridized the ICP optimization and active shape model (ASM) fitting to enable the automatic registration of 3D facial surfaces [52]. They demonstrated that for the ten studied landmarks, the estimated positions using the DSM method have relatively small RMS errors (~3*mm*) from the manual annotations. Our method achieved much lower landmark RMS errors, 1.7~1.8 *mm* on average, for a bigger number (17) of landmarks (table 1). If the less salient landmarks, such as the earlobe tips, are excluded from the analysis, the errors will decrease further. The gain in accuracy may be partially attributed to the higher image resolution of our data (~30,000 vertices per surface on average) compared to the previous work (~10,000 vertices per surface). Nonetheless we believe that the use of both shape and texture information played a key role in improving the landmark localization accuracy. We found that the positions of some salient landmarks such as the eye corners are ambiguous even manually when the texture is removed. Texture gives rich information about the facial anatomical layout, such as the boundaries of different skin/tissue types. In fact, texture is almost the only information source for pattern recognition in 2D images and has been shown to give good performance. We projected both the shape and texture data into the 2D space, where the well-established PCA algorithm was used to detect the key landmarks. We also made use of the texture information for detecting certain other landmarks. This design seems to improve the accuracy and robustness of our method significantly. Furthermore, due to the use of simple and optimized algorithms, the landmark annotation is also very efficient and does not require large amounts of memory. Hundreds of surfaces can be annotated within several minutes on a standard Windows PC. It should be noted that the aim of this study is not to propose a general scheme of landmark recognition on 3D surfaces. Rather, it gives a pragmatic solution for automatically annotating a fixed set of salient landmarks on the human face. Nonetheless, our new idea of combining both the shape and texture information in the PCA framework may be extended to improve general feature recognition on 3D surfaces.

In this work, we also proposed a new protocol for the TPS-based registration, whereby the TPS warping was only applied to the reference face while the sample faces remained undeformed and thus avoided the step of inverse TPS warping, thereby further increasing the efficiency of our method. On the other hand, the robustness of the registration is well retained, as can be seen from the results. To address the surface mesh irregularities such as the caveats and uneven vertex densities, we proposed a spherical sampling step to re-mesh the reference surface. This reference surface, when used in the registration, resulted in similarly well structured meshes in the sample surfaces as well. The registration quality can be further improved by replacing the original reference face with a sample-wide average face, followed by a second registration. This can further reduce the dependency of results on a specific reference.

It is interesting to note that both the automatic landmark annotation and the TPS based registration steps work equally well for two different ethnicities, namely Han Chinese and Uyghur, in spite of the fact that they are substantially different in both genetic background and facial appearance. Han Chinese are representative of East Asian populations while Uyghur is an ancient admixture population whose ancestries came from both East Asians and Caucasians (European people) [53]. As a result, Uyghur participants exhibited many Caucasian facial features such as sunken eyes and high nose ridge, etc (figure 8). This method was also tested on other ethnic groups and showed consistent robustness (data not shown). Such ethnicity independency is very important when this method is used to study the cross population facial morphological variations in humans. Limited by the sample collection, we are not able to analyze the sensitivity of our parameters with respect to very different age groups, or atypical faces caused by diseases or obesity. To assist studies of other facial morphological variations such as face deformation diseases, face growth, and aging, this method has to be extended and tested for the specific datasets. Furthermore, the anatomic correspondence can be further improved by including additional features such as the eyebrows, eyelid lines, and lip lines as landmarks. These features may provide discrimination power towards different facial expressions. Nonetheless, we believe the approach we describe here can provide a good basis for 3D face dense registration in general.

In summary, this study proposes a new scheme to build accurate and robust anatomical correspondence across dense surfaces of 3D facial images; and it can be implemented into a fully automatic and efficient registration method. This method enables high-throughput capture and analysis of the wide ranging and yet fine detailed variations within human facial morphology. Such comprehensive and high resolution phenotypic data should be valuable in anthropological, disease diagnosis, and forensic studies of human facial morphology.

# Acknowledgement

Dr. Mark Stoneking from the Max Planck Institute for Evolutionary Anthropology has made valuable contributions to this study and provided proof-reading of the manuscript.

## Appendix I:

Denoting a facial mesh composed of $N$ points by $\mathbf{F} = \{\mathbf{p}_i\}$ for $i = 1, \ldots, N$. Suppose $\mathbf{S}$ is the set of $M$ points that are within distance $R$ around the point $\mathbf{p}_j (1 \leq j \leq N)$. The best fit sphere $\mathbf{T}$ around $\mathbf{p}_j$ is therefore determined by two parameters, namely the center $\mathbf{O} = (a, b, c)$ and radius $r$. The squared distance from each point $\mathbf{p}_k = (x_k, y_k, z_k)$, $1 \leq k \leq M$ in $\mathbf{S}$ to the surface of $\mathbf{T}$ is defined by

$$d^2(\mathbf{p}_k, \mathbf{T}) = \left| (x_k - a)^2 + (y_k - b)^2 + (z_k - c)^2 - r^2 \right| \tag{s1}$$

Let us denote $d^2(\mathbf{p}_k, \mathbf{T})$ as $\varepsilon_k$, then the above equation can be expressed as

$$\varepsilon_k = \left| (x_k^2 + y_k^2 + z_k^2) - (2ax_k + 2by_k + 2cz_k) + (a^2 + b^2 + c^2 - r^2) \right| \tag{s2}$$

And the square distance vector is

$$\boldsymbol{\varepsilon} = [\varepsilon_1, \ldots, \varepsilon_M]^T \tag{s3}$$

Our goal is to minimize the following error function

$$E = \sum_{k=1}^{M} \varepsilon_k^2 = \boldsymbol{\varepsilon}^T \boldsymbol{\varepsilon} = (\mathbf{A} - \mathbf{B}^T \mathbf{W})^T (\mathbf{A} - \mathbf{B}^T \mathbf{W}) \tag{s4}$$

Where

$$\mathbf{A} = \begin{bmatrix} x_1^2 + y_1^2 + z_1^2 \\ \vdots \\ x_k^2 + y_k^2 + z_k^2 \end{bmatrix}$$

$$\mathbf{W} = [2a, 2b, 2c, r^2 - a^2 + b^2 + c^2]^T$$

$$\mathbf{B} = \begin{bmatrix} x_1 & \cdots & x_k \\ y_1 & \cdots & y_k \\ z_1 & \cdots & z_k \\ 1 & \cdots & 1 \end{bmatrix}$$

This is a simple least squares problem. The solution is $\mathbf{W} = (\mathbf{BB}^T)^{-1} \mathbf{BA}$, and the radius is $r = \sqrt{W(4) + (W(1)^2 + W(2)^2 + W(3)^2)/4}$. The radius $r$ is a key measurement for nose-tip recognition. In order to assess how close the point set **S** matches the sphere **T**, we introduce another measurement: the mean fitting residual, defined as $e = \sqrt{E/M}$. The smaller $e$ is, the better **S** fits to a sphere.

## Appendix II:

The vector P is defined as in the main text equation 2. Denote the mean of the $P$ vectors across the training set as $\bar{P}_t$, the covariance matrix is calculated as

$$\mathbf{C} = \frac{1}{m-1} \sum_{i=1}^{m} (P_{t,m} - \bar{P}_t)(P_{t,m} - \bar{P}_t)^T \tag{s5}$$

The eigen space **U** is then constructed by the eigenvectors $u_i$ such that

$$\mathbf{C} u_i = \lambda_i u_i \tag{s6}$$

where $\lambda_i$ is the $i$th largest eigen value of **C**. And **U** is given by $\mathbf{U} = [u_1, u_2, ..., u_k]$. Here $k$ is the actual number of eigen vectors to be used, which is set to 16 in our case. **U** therefore defines an eigen space where the sample $P$ patches can be evaluated for similarity.

For a sample face, every point in the 2D grid is given a 21$mm \times$21$mm$ patch and a sample patch vector $P_s$ is similarly derived following equation (2). $P_s$ is then subtracted by $\bar{P}_t$ and projected into the eigen space **U** to give the weight vector

$$w = \mathbf{U}^T (P_s - \bar{P}_t) = [\omega_1, \omega_2, \cdots, \omega_k]^T \tag{s7}$$

$P_s$ can be reconstructed using $w$ as $P_s' = \bar{P}_t + w\mathbf{U} = \bar{P}_t + \mathbf{U}^T(P_s - \bar{P}_t)\mathbf{U}$. The

reconstruction error can be described as

$$e = (P_s - P_s')^T (P_s - P_s') \tag{s8}$$

A valid landmark point should lie close to the origin point in the **U** space; we therefore use only points satisfying $-3\sqrt{\lambda_i} \leq \omega_i \leq 3\sqrt{\lambda_i}$, where $\lambda_i$ is the variance along $\omega_i$ across the training set. We also calculate the Mahalanobis distance from $P_s'$ to $\overline{P_t}$.

$$d = \sqrt{\sum_{i=1}^{k} \frac{\omega_i^2}{\lambda_i}} \tag{s9}$$

which can be another indicator of pattern similarity.